\documentclass[runningheads]{llncs}
\usepackage{graphicx}

\usepackage{multirow}
\usepackage{multicol}
\usepackage{algpseudocode}
\usepackage{graphicx}
\usepackage{textcomp}
\usepackage{xcolor}
\usepackage{multirow}
\usepackage{booktabs}
\usepackage{subfigure}
\usepackage{hyperref}
\usepackage{amsmath,amssymb,amsfonts}
\usepackage{algorithm}
\usepackage{algorithmicx}

\usepackage[T1]{fontenc}
\def\doi#1{\href{https://doi.org/\detokenize{#1}}{\url{https://doi.org/\detokenize{#1}}}}
\usepackage{graphicx}
\usepackage{listings}
\begin{document}
\title{Learning Robust Representation through Graph Adversarial Contrastive Learning}
\titlerunning{Graph Adversarial Contrastive Learning}
%
\author{Jiayan Guo \inst{1,\star \dag} \and
Shangyang Li\inst{2,\star \dag} \and
Yue Zhao\inst{3,\star } \and
Yan Zhang \inst{1}
}

\renewcommand{\thefootnote}{}
\footnotetext{$\star$ $\quad$ Equal\; contribution\\
$\dag$ $\quad$ Corresponding\; authors}

%
\authorrunning{Guo. et al.}
%
\institute{
School of Artificial Intelligence, Peking University, Beijing, China\\
\email{\{guojiayan, zhyzhy001\}@pku.edu.cn}\and
Peking-Tsinghua Center for Life Sciences, IDG/McGovern Institute for Brain
Research, Academy for Advanced Interdisciplinary Studies, Peking University, Beijing, China\\
\email{syli@pku.edu.cn} \and
Academy for Advanced Interdisciplinary Studies, Peking University, Beijing, China \\
\email{zhaoyue@stu.pku.edu.cn}
}
\maketitle              
\begin{abstract}
Existing studies show that node representations generated by graph neural networks (GNNs) are vulnerable to adversarial attacks, such as unnoticeable perturbations of adjacent matrix and node features. Thus, it is requisite to learn robust representations in graph neural networks. To improve the robustness of graph representation learning, we propose a novel \textbf{Graph} \textbf{A}dversarial \textbf{C}ontrastive \textbf{L}earning framework (GraphACL) by introducing adversarial augmentations into graph self-supervised learning.
In this framework, we maximize the mutual information between local and global representations of a perturbed graph and its adversarial augmentations, where the adversarial graphs can be generated in either supervised or unsupervised approaches.
Based on the Information Bottleneck Principle, we theoretically prove that our method could obtain a much tighter bound, thus improving the robustness of graph representation learning. Empirically, we evaluate several methods on a range of node classification benchmarks and the results demonstrate GraphACL could achieve comparable accuracy over previous supervised methods.

\end{abstract}
\section{Introduction}
Graph neural networks (GNNs) have enabled significant advances on graph-structured data~\cite{kipf2016semi,velivckovic2017graph} and are widely used in many applications like node classification, graph classification, and recommendation systems. However, existing works show that they are vulnerable towards adversarial attacks~\cite{10.1145/3219819.3220078,zugner_adversarial_2019,10.1145/3394520} like unnoticeable perturbations, which is still a critical challenge in employing GNNs in safety-critical applications.

Albeit various studies have been proposed to ensure the robustness of the graph neural networks against adversarial attacks~\cite{tang2020transferring,zhu2019robust,10.1145/3394486.3403049,ijcai2019-669}, the significance of adversarial augmentations has been ignored, especially under unsupervised learning setting. Recently, self-supervised learning has achieved remarkable performances on graph-structured data, like DGI~\cite{velivckovic2018deep}, GraphCL~\cite{NEURIPS2020_3fe23034}, etc. These works use pairs of augmentations on unlabeled graphs to define a classification task for pretext learning of graph representations. Also, GraphCL~\cite{NEURIPS2020_3fe23034} has found that contrastive learning with randomly generated graph augmentations can somehow increase the robustness; however, we argue that such randomly generated samples are not the optimal choice to achieve the robustness of representations and adversarial augmentations can perform provably much better.

Thereby, we present a novel adversarial self-supervised learning framework to learn robust graph representations.
We introduce adversarial samples into the input.
Primairly, both supervised and unsupervised approaches can be used to generate adversarial samples. For example, Metattack~\cite{zugner_adversarial_2019}, a supervised adversarial attack method, can be directly applied. 
Besides, we further propose an unsupervised method to generated adversarial graphs, which uses unsupervised contrastive loss as the target of Metattack to generete adversarial samples. After generating perturbed graphs, we maximize the similarity between representations of the clean graph and the adversarial attacked graph to suppress distortions caused by adversarial perturbations. This will result in representations that are robust against adversarial attacks.

We refer to this novel adversarial self-supervised graph representation learning method as \textbf{Graph} \textbf{A}dversarial \textbf{C}ontrastive \textbf{L}earning (GraphACL). To the best of our knowledge, this is the first attempt to use adversarial samples to increase the robustness of graph representations based on contrastive learning. We also build a theoretical framework to analyze the robustness of graph contrastive learning based on the Information Bottleneck Principle. To verify the effectiveness of GraphACL, we conduct experiments on public academic dataset, Cora, Citeseer, Pubmed under both targeted attack~(i.e., Netattack) and global attack~(i.e., Metattack). Experimental results suggest that GraphACL outperforms DGI and other baselines significantly, thus proving our method can learn robust representations under various graph adversarial attacks.

\begin{figure*}[hbtp]
  \centering
  \includegraphics[width=\textwidth]{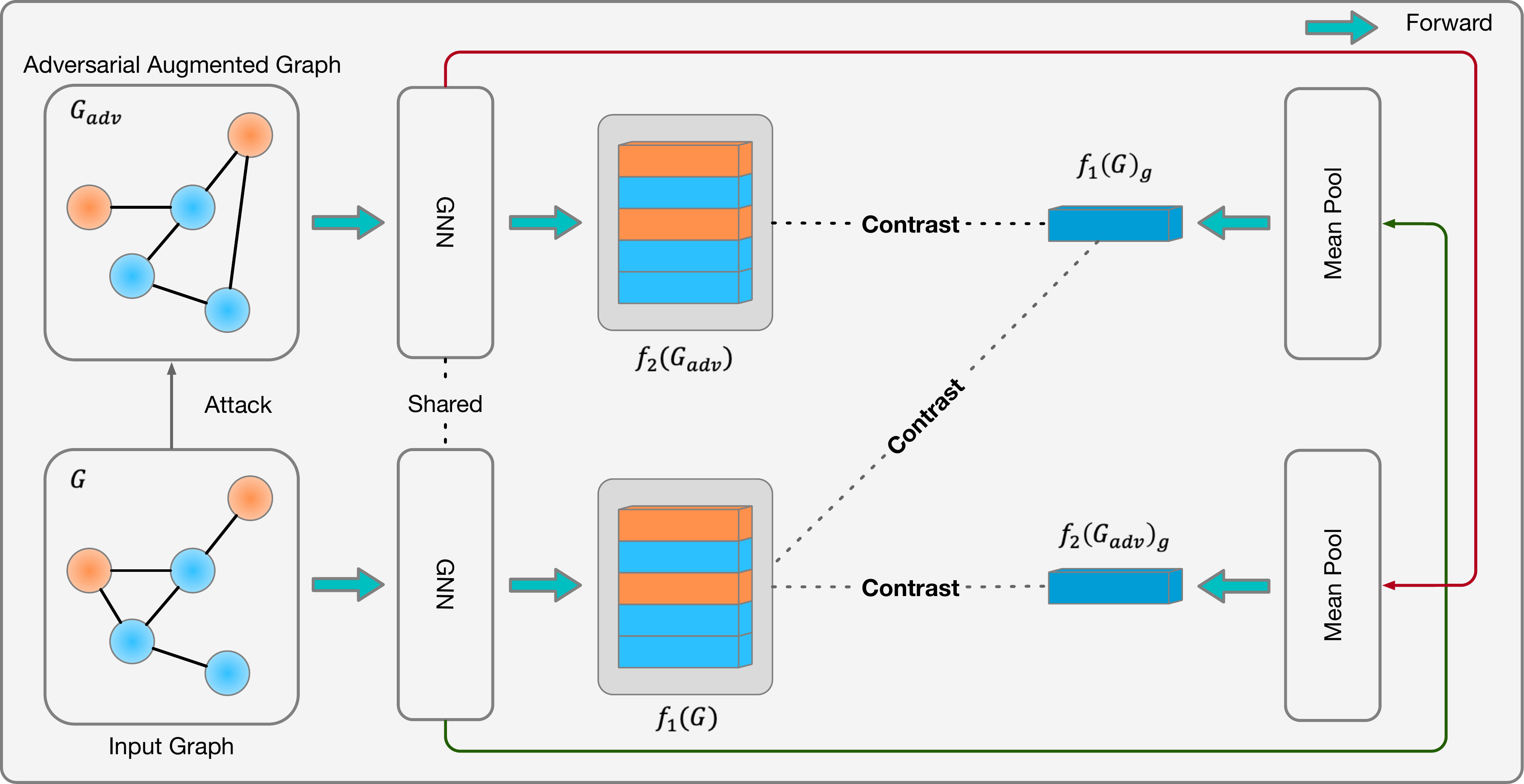}
  \caption{Graph Adversarial Contrasive Learning Framework}
  \label{fig:framework}
\end{figure*}

In summary, our contributions are as follows:

\begin{itemize}
    \item We propose GraphACL, a general framework to use self-supervised graph contrastive learning with adversarial samples to learn robust graph representations.
    \item We theoretically prove that our method could improve the robustness of graph representation learning from the perspective of information theory.
    \item We present an unsupervised graph adversarial attack method that use meta-gradient to poison the graph structure to maximize the contrastive loss between clean and perturbed graphs.
    \item We conduct extensive experiments to demonstrate the effectiveness of our proposed GraphACL under various types of adversarial attacks, which indicates that GraphACL can significantly improve the performance of previous methods both in evasive and poisoning settings.
\end{itemize}

\section{Methodologies}

\subsection{Graph Adversarial Attack}

In this subsection, we will formulate the classic optimization problem of graph adversarial attack.
Let $G=(A,X)$ be an attribute graph with adjacency matrix $A\in\{0,1\}^{N\times N}$ and attribute matrix $X\in\mathbb{R}^{N\times D}$, where $N$ is the number of nodes and $D$ is the dimension of the node feature vector. Considering a semi-supervised node classification task, where labels of the nodes $\mathcal{V}_L\in \mathcal{V}$ are given. Each node is assigned as one class in $\mathcal{C}\in\{c_1,...,c_k\}$. The goal of adversarial attack can be mathematically formulated as a bilevel optimization problem

\begin{equation}
\begin{aligned}
    &\mathop{\text{max}}_{{G}_{attack}\in\Phi(G)}\mathcal{L}(f_{\theta^*}({G}_{attack}))\\  & s.t. \ \ \theta^*=\mathop{arg min}_\theta \mathcal{L}(f_\theta({G}))
\end{aligned}
\end{equation}

\noindent where $\Phi(G)$ is the space of perturbation on the input graph, $\mathcal{L}$ is the cross entropy by default and $f_\theta(\cdot)$ is the surrogate model.

Based on whether to re-train the model on the attacked graph, the attack type is categorized by poisoning attack and evasive attack. Poisoning attack requires re-training while evasive attack does not.

\subsection{Graph Adversarial Contrastive Learning Framework}

As illustrated in Figure~\ref{fig:framework}, we now present our framework to learn robust representations via adversarial contrastive training. 
Firstly, we conduct adversarial generation on the perturbed graph. Then, we use the input graph and an adversarial augmented graph as different views of the same graph. A shared encoder like GCN encodes multi-views of the graph and then outputs respective local representations $f_1(G)$ and $f_2(G_{adv})$, where $G$ is the input graph and $G_{adv}$ is the adversarial augmentation.  $f_1(\cdot)$ and $f_2(\cdot)$ are encoders that can be the same or different with their unshared projection layers. Crossed local-global information maximization is implemented by maximizing the information between local representations of the input graph and global representations of the adversarial graph, vice versa. The GraphACL framework is modified on DGI framework by additionally introducing an adversarial augmented view of the
input graph. The other omitted settings are the same with DGI, and negative samples are also used. Therefore, the improvement of GraphACL over DGI is of our concern.

\begin{figure*}[htbp]
  \centering
  \includegraphics[width=.6\textwidth]{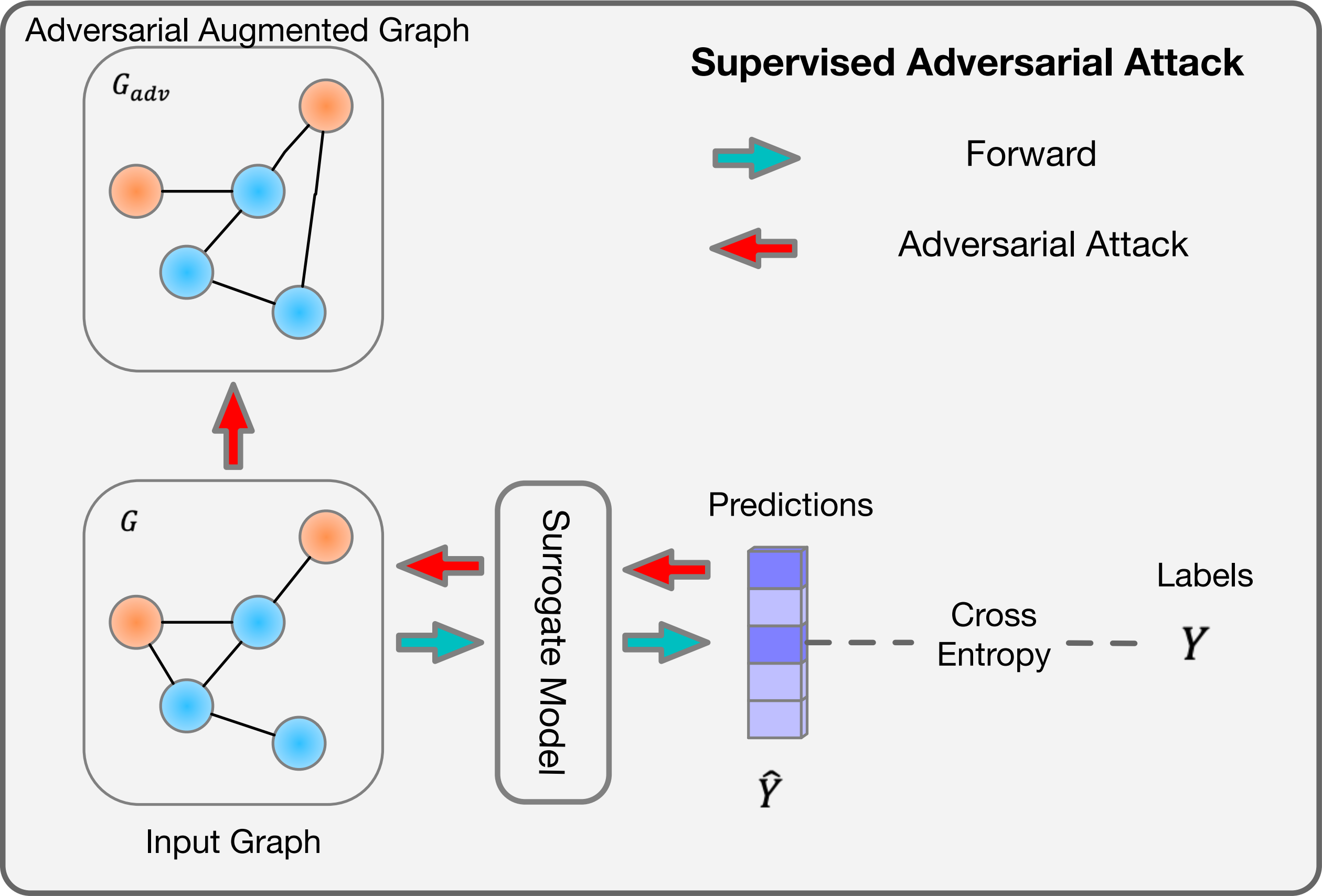}
  \caption{Generation of graph adversarial augmentations under supervised loss.}
  \label{fig:attack_su}
\end{figure*}

When several labels are known, adversarial augmentation ${G}_{adv}$ could be obtained by supervised generation method.
The process is illustrated in Figure \ref{fig:attack_su} Then
we use the contrastive learning objective to maximize the similarity between input examples ${G}$ and their instance-wise adversarial augmentation ${G}_{adv}$. Then we can formulate our Graph Adversarial Contrastive Learning objective as follow:
\begin{small}
\begin{equation}
\begin{aligned}
\label{eq:sup}
\mathcal{L}_{GACL}^{sup} = &\mathop{\min}_{f_1,f_2}  (L_{cl}^{self}(  f_1({G}), f_1({G})_{global})  \\
&+ \alpha L_{cl}^{adv}(  f_1({G}), f_2({G}_{adv})_{global}) \\
&+ \beta L_{cl}^{adv}(  f_1({G})_{global}, f_2({G}_{adv}))),
\end{aligned}
\end{equation}
\end{small}where $L_{cl}$ is contrastive loss that is negative mutual information essentially and $\alpha$ balances between contrastive loss $L_{cl}^{self}$ and $L_{cl}^{adv}$. Similar to DGI \cite{velivckovic2018deep}, $f(\cdot)_{global}$ is the global representation of the whole graph.

\begin{figure*}[htbp]
  \centering
  \includegraphics[width=.8\textwidth]{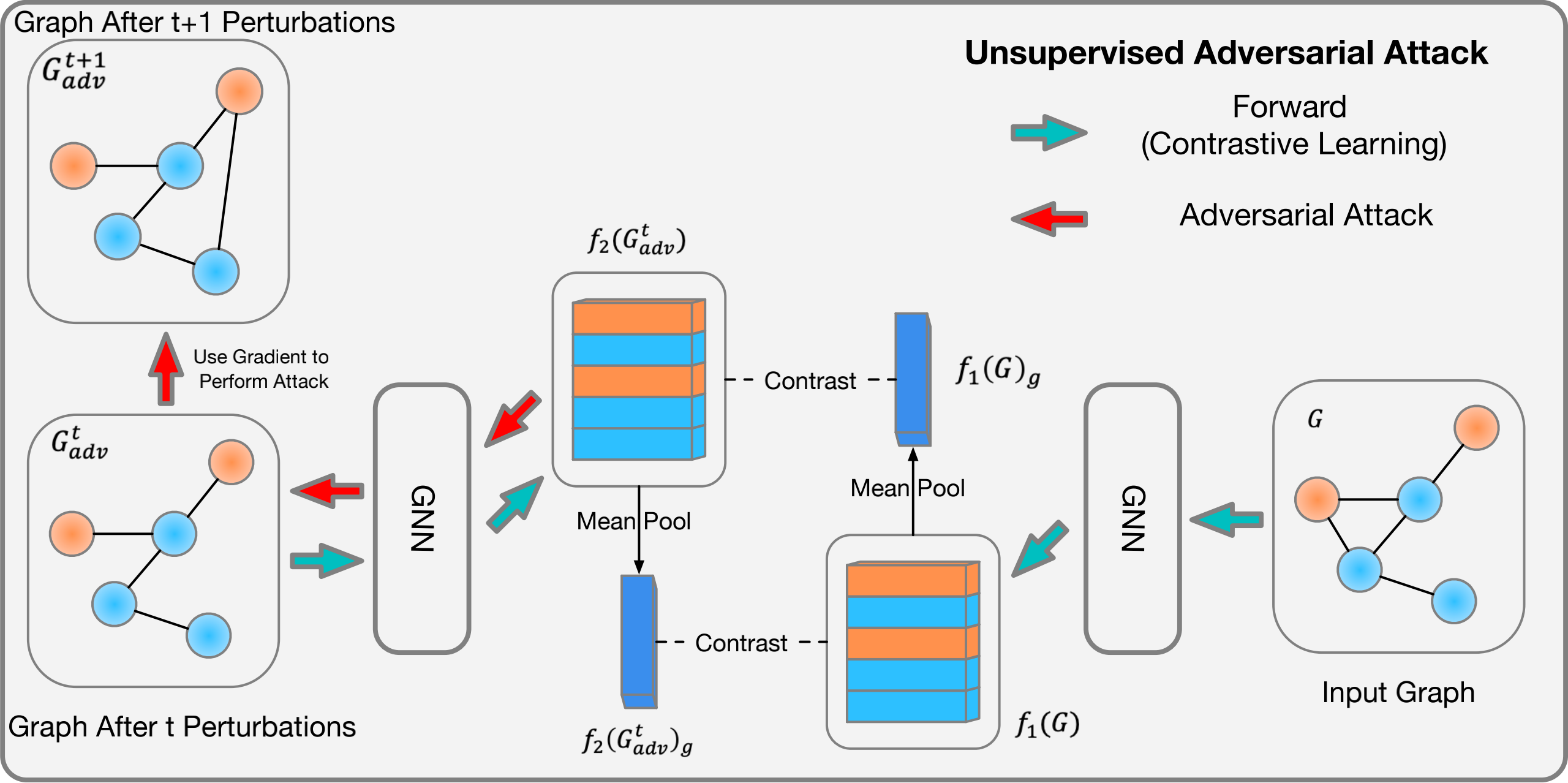}
  \caption{Generation of graph adversarial augmentations under unsupervised loss.}
  \label{fig:attack_unsu}
\end{figure*}

If no label information is given, the unsupervised adversarial training strategy of Graph Adversarial Contrastive Learning objective can be formulated as:
\begin{small}
\begin{equation}
\begin{aligned}
\label{eq:unsup}
\mathcal{L}_{GACL}^{unsup} = &\mathop{\min}_{f_1,f_2} \mathop{\max}_{g} (L_{cl}^{self}(  f_1({G}), f_1({G})_{global})\\
&+ \alpha L_{cl}^{adv}(  f_1({G}), f_2(g({G}))_{global}) \\
&+ \beta L_{cl}^{adv}(  f_1({G})_{global}, f_2(g({G})))),\\
\end{aligned}
\end{equation}
\end{small}
where $g(\cdot)$ is adversarial  samples generation function. The process is illustrated in Figure \ref{fig:attack_unsu}

The detailed procedure is presented as \textbf{Algorithm \ref{al:1}}. Different ways of generating adversarial augmentations are formulated in the next section of theoretical analysis.

 \begin{algorithm}[h]
  \caption{The Procedure of One Iteration in GraphACL}
  \label{al:1}
  \begin{algorithmic}[1]
    \Require
      Input Graph $G=(A,X)$;
    \Ensure
      $f_1(\cdot)$, $f_2(\cdot)$: graph encoders;

    \If{use supervised adversarial augmentation}
    \State generate adversarial graph $\hat{G}$ based on Eq.(\ref{adv-obj});
    \Else
    \State generate adversarial graph $\hat{G}$ based on Eq.(\ref{adv-obj-un});
    \EndIf

    \State Generate node representations of input graph $f_1(G)$;
    \State Generate node representations of adversarial augmented graph $f_2(G_{adv})$;
    \State Generate global representation of input graph $f_1(G)_g$ by mean pooling;
    \State Generate global representation of adversarial augmented graph $f_2(G_{adv})_g$ by mean pooling;
    \State Compute contrastive loss by Eq.(\ref{eq:sup}) or Eq.(\ref{eq:unsup});
    \State Back propagate gradients and update $f_1(\cdot)$ and $f_2(\cdot)$;\\

    \Return $f_1(\cdot)$ and $f_2(\cdot)$;
  \end{algorithmic}
\end{algorithm}

\section{Theoretical Analysis on Graph Adversarial Contrastive Learning}
In this section, we first formulate the Information Bottleneck Principle in graph self-supervised learning and achieve the related objective function. Then, we illustrate the generation of adversarial augmentations in Figure~\ref{fig:attacks}. Finally, we derive desirable lower bounds of the information bottleneck and formulate the objective function according to supervised and unsupervised adversarial augmentations.

\subsection{Information Bottleneck Principle for Graph Self-supervised Learning}
The Information Bottleneck (IB) \cite{tishby2000the,tishby2015deep} provides an essential principle for representation learning from the perspective of information theory, which is an optimal representation need to contain minimal yet sufficient information for downstream tasks. It encourages the representation to involve as much information about the target as possible to obtain high prediction accuracy, and discard redundant information that is irrelevant to the target. In graph representation learning, each graph ${G}(A, X)$ contains information of both the graph structure $A\in\mathbb{R}^{N\times N}$ and node features $X\in\mathbb{R}^{N\times d}$. Applying IB to graph self-supervised learning, we desire to learn an optimal graph representation $Z$, which is informative about the original graph ${G} \in {{\mathcal{G}}}$ , but invariant to its augmentations $\widehat{G} \in {\widehat{\mathcal{G}}}$. This principle can be formulated as follows:
\begin{equation}\label{ibloss}
{\mathcal{L}}_{\mathrm{IB}} \triangleq \beta I(Z, \widehat{\mathcal{G}})- I\left(Z, \mathcal{G}\right),
\end{equation}
where $I(\cdot,\cdot)$ denotes mutual information between variables and $\beta>0$ is a hyper-parameter to control the trade-off between preserving information and being invariant to distortions.

We use $\widehat{G}$ to represent different views of the corresponding graph $G$. For the first term of ${\mathcal{L}}_{\mathrm{IB}}$, we utilize an upper bound proved in \cite{cheng2020club} to derive a tractable bound of the mutual information between $Z$ and $\widehat{G}$:
\begin{equation}\label{term2}
\begin{aligned}
I(Z, \widehat{\mathcal{G}}) \leq & \sum_{z \in {Z}} \sum_{\widehat{G} \in {\widehat{\mathcal{G}}}} p(\widehat{G}, z) \log (p(z \mid \widehat{G})) \\
& -\sum_{z \in {Z}} \sum_{\widehat{G} \in {\widehat{\mathcal{G}}}} p(\widehat{G}) p(z) \log (p(z \mid \widehat{G})),
\end{aligned}
\end{equation}

Also, the mutual information between $Z$ and $\mathcal{G}$ can be written as
\begin{equation}
\begin{aligned}
I(Z , \mathcal{G}) = & \sum_{z \in {Z}} \sum_{G \in {\mathcal{G}}} p(z, G) \log \frac{p(z, G)}{p(z) p(G)} \\
= & \sum_{z \in {Z}} \sum_{G \in {\mathcal{G}}} p(z, G) \log p(G | z) + H(\mathcal{G}). \\
\end{aligned}
\end{equation}
The entropy term $H(\mathcal{G})$ could be dropped which results in
\begin{equation}\label{term1}
\begin{aligned}
I(Z, \mathcal{G}) & \geq  \sum_{z \in {Z}} \sum_{G \in {\mathcal{G}}} p(z, G) \log p(G | z).\\
\end{aligned}
\end{equation}

By combining Eq.(\ref{ibloss}), Eq.(\ref{term2}) and Eq.(\ref{term1}), we can minimize the upper bound of IB by:

\begin{equation}
\begin{aligned}
\hat{\mathcal{L}}_{\mathrm{IB}} =
&\beta \frac{1}{NM} \sum_{i=1}^{N}  \sum_{j=1}^{M} [\log p(z_{j}^{i} \mid \widehat{G}_{j}^{i})\\
&- \frac{1}{M} \sum_{k=1}^{M} \log  p(z_{k}^{i} \mid \widehat{G}_{j}^{i})] \\
& - \frac{ 1}{NM} \sum_{i=1}^{N} \sum_{j=1}^{M} \log p(G^{i} \mid z_{j}^{i}),\\
\end{aligned}
\label{loss}
\end{equation}
where $N$ is the number of original graphs, $M$ is the number of augmentations of each original input graph. In graph contrastive learning, $p(z_{j}^{i} \mid \widehat{G}_{j}^{i})$ can be viewed as an encoder $f_{\theta}$: $\widehat{\mathcal{G}} \rightarrow Z$.

We assume
$p(z_{k}^{i} \mid z_{j}^{i}, \widehat{G}_{j}^{i}) = p(z_{k}^{i} \mid z_{j}^{i})$,
which means the representation $z_{k}^{i}$ of an augmented graph cannot depend directly on another augmented graph $\widehat{G}_{j}^{i}$. Also, since the function $f_{\theta}$ is deterministic, we have
\begin{equation}
\begin{aligned}
p(z_{k}^{i} \mid \widehat{G}_{j}^{i}) = & \sum p(z_{k}^{i} \mid z_{j}^{i}, \widehat{G}_{j}^{i})p(z_{j}^{i} \mid \widehat{G}_{j}^{i}) \\
= & p(z_{k}^{i} \mid z_{j}^{i})p(z_{j}^{i} \mid \widehat{G}_{j}^{i}) = p(z_{k}^{i} \mid z_{j}^{i})
\end{aligned}
\end{equation}
Further, Eq.(\ref{loss}) can be written as:
\begin{equation}
\begin{aligned}
\hat{\mathcal{L}}_{\mathrm{IB}} =& \beta \frac{ 1}{NM} \sum_{i=1}^{N} \sum_{j=1}^{M} [ - \frac{1}{M}\sum_{k=1}^{M} \log p(z_{k}^{i} \mid z_{j}^{i})] \\
& - \frac{ 1}{NM} \sum_{i=1}^{N} \sum_{j=1}^{M} \log p(G^{i} \mid z_{j}^{i}),
\end{aligned}
\label{loss1}
\end{equation}
where $p(z_{k}^{i} | z_{j}^{i})$ could be viewed as a similarity measurement between representations of different augmentations. Eventually, we formulate the problem with IB and obtain a general objective function of graph self-supervised learning as Eq.(\ref{loss1}). Intuitively, the objective function motivates GNN to increase the averaged similarity of representations between different augmentations, thus making the learned representations invariant and robust to various different views.

\subsection{Generation of Supervised Graph Adversarial Augmentations}\label{loss-sup}
The generation of supervised graph adversarial augmentation is schematically shown in Figure~\ref{fig:attack_su}, which utilizes previous graph adversarial attack methods like Metattack~\cite{zugner_adversarial_2019}. Primarily, a surrogate model is applied to the perturbed graph to generate predictions. Then the supervised loss is computed by cross entropy. Finally, we use the gradient to modify the structure of the original graph to generate adversarial samples.

Suppose $G$ is the original graph and its node labels are $Y$, we consider a softmax regression layer between $Z$ and $Y$. The posterior class probabilities can be written as:
\begin{equation}
\begin{aligned}
P_{Y \mid Z}(y \mid z) &=\frac{e^{w_{y}^{T} z}}{\sum_{k} e^{w_{k}^{T} z}}\\
&= \frac{e^{w_{y}^{T} f_{\theta}({G})}}{\sum_{k} e^{w_{k}^{T} f_{\theta}({{G}})}}\\
\end{aligned}
\end{equation}
where $\mathcal{W}=\left\{w_{y}\right\}_{y=1}^{k}$ is the vector of classification parameters for class $y$ and $\theta$ is the parameter of the encoder $f_\theta(\cdot)$, which are learned by minimizing the cross-entropy loss
\begin{equation}
L_{c e}( {G},Y ; \mathcal{W}, \theta)=-\log \frac{e^{w_{y}^{T} f_{\theta}({G})}}{\sum_{k} e^{w_{k}^{T} f_{\theta}({G})}}.
\end{equation}

Given the learned encoder and classifier, an optimal perturbation for $G$ is generated by maximizing the cross-entropy loss:
\begin{equation} \label{adv-obj}
\begin{aligned}
&G_{adv*}=\arg \max _{G_{adv}} L_{c e}(G_{adv}, y ; \mathcal{W}, \theta) \\
&\quad \text { s.t. }  G_{adv} \in \Phi\left(G\right),
\end{aligned}
\end{equation}
where $\Phi\left(G\right)$ means the space of perturbation on the original graph. Then we can further formulate a constrained optimization problem as following
\begin{equation}
\begin{aligned}
& G_{adv} \in \Phi\left(X\right) \\
\text { s.t. } & \mathcal{Q}\left(G, G_{adv}\right)<\epsilon
\end{aligned}
\end{equation}
where $\mathcal{Q}(\cdot)$ represents a distance measurement function, $\epsilon$ is a parameter for imperceptible perturbation evaluation.

\begin{equation}
\begin{split}
&\ \frac{1}{K} \sum_{k=1}^{M} \log p[f_{\theta}(\widehat{G}) \mid f_{\theta}(G)]\\
= &\ \frac{1}{K} \sum_{k=1}^{M} \log{ \sum p[f_{\theta}(\widehat{G}) \mid f_{\theta}(G), y)p(y \mid f_{\theta}(G)]}\\
= &\ \frac{1}{K} \sum_{k=1}^{M} \log  \sum p[f_{\theta}(\widehat{G}) \mid y)p(y \mid f_{\theta}(G))]\\
> &\ \log [\sum p( f_{\theta}({G}_{adv*}) \mid y)p(y \mid f_{\theta}(G))]\\
= &\ \log p[ f_{\theta}({G}_{adv*}) \mid f_{\theta}(G)],
\end{split}
\end{equation}
Since similarity between clean graph representations and adversarial augmentation representations becomes a lower bound of the averaged similarity between representations on the original graph and all augmentations.

\subsection{Generation of Unsupervised Graph Adversarial Augmentations}\label{loss-unsup}

The unsupervised graph adversarial augmentation generation is schematically in Figure~\ref{fig:attack_unsu}.
Graph $i$ is mapped into an example pair $(\widehat{G}^{i}_{k}, \widehat{G}^{i}_{j})$. Graph contrastive learning is performed through maximizing the agreement between an positive pair. Eq.(\ref{loss1}) tells us that if we want to get a more robust representation, we need to increase $\frac{1}{K}\sum_{k=1}^{M} \log p(z_{k}^{i} | z_{j}^{i})$. Similar to supervised situation, there is a lower bound $\frac{1}{K} \sum_{k=1}^{M} \log p[f_{\theta}(\widehat{G}^{i}_{k}) | f_{\theta}(\widehat{G}^{i}_{j}) ] > \log p[ f_{\theta}({G}_{adv*}^{i})| f_{\theta}(\widehat{G}^{i}_{j})]$.

The choice of ${G}_{adv*}$ could be formulate as a two-stage alternative optimization problem: one is self-supervised learning, the other is adversarial attack or generation of adversarial augmentations. A generative function $g(\cdot)$ is introduced to denote the generation of adversarial samples ${G}_{adv*}$. For example, the generation function $g(\cdot)$ can be the same as it in Metattack. In the first stage, adversarial samples can be generated by ${G}_{adv} = g({G})$, which is further considered as an augmentation or a different view of the perturbed ${G}$. Hence, self-supervised learning is conducted to maximize the mutual information between different views by optimizing corresponding encoders $f_1$ and $f_2$.
Then, given the encoders, we can optimize $g(\cdot)$ by using adversarial attacks to minimize self-supervised loss and obtain a new adversarial graph.
Still using Metattack as an example, $g(\cdot)$ is optimized by attacking the gradient of the self-supervised loss.

Finally, we formulate the two-stage of unsupervised training strategy -- adversarial attack and self-supervised learning as an underlying min-max objective function in the following:
\begin{equation}
 \min _{g} \max _{f_{1}, f_{2}}  I(f_1({G}), f_2(g({G})))
\end{equation}

In practice, as shown in Figure~\ref{fig:attack_unsu}, the adversarial unsupervised training strategy is modified on DGI framework:
\begin{equation}
\label{adv-obj-un}
\begin{aligned}
 \min _{g} \max _{f_{1}, f_{2}}  \{ &I[f_1({G})_{global}, f_2(g({G}))] \\
 &+ I[f_1({G}), f_2(g({G}))_{global}] \}.
\end{aligned}
\end{equation}

\section{Experiments}

\subsection{Experimental Settings}
To evaluate the robustness of different models against adversarial attacks, we conduct experiments on the following benchmarks with Netattack \cite{pmlr-v97-bojchevski19a} and Metattack \cite{zugner_adversarial_2019}, where adversarial augmentations are generated by supervised or unsupervised contrastive loss, respectively. Netattack works based on boolean features; therefore, the features in each dataset are preprocessed to be 0 or 1. We follow the experimental settings in Netattack \cite{pmlr-v97-bojchevski19a,NEURIPS2020_3fe23034,wu2020graph} exactly: We test the classification accuracy of the 40 selected target nodes: 10 nodes with the highest margin of classification, which is most likely to be classified correctly; 10 nodes with the lowest margin of classification but still classified correctly, which may be easily attacked; 20 other random nodes. Each perturbation denotes a filp on a boolean feature or a modification on an edge related to the node. Robustness experiments are evaluated on a clean graph and corrupted graphs with a number of perturbations from 1 to 4. For Metattack, we use the standard Metattack setting with a perturbation rate of 0.05 and 0.2 on clean graphs to generate modified graphs. Then we test the node classification accuracy on the modified graph. In this experiment, the adversarial augmentations are generated by our proposed unsupervised contrastive loss.

We evaluate two types of robustness tasks, including evasive and poisoning. We include the baselines such as GCN, RGCN, GAT; the results are cited from \cite{wu2020graph}. Previous work has not included the pre-trained models in attack experiments, and DGI is now considered to be compared with since we desire to evaluate the impact of adversarial augmentations. GIB \cite{wu2020graph} is one of the previous SOTA on these experiments; however, it is not related to our comparison on whether to use adversarial augmentations nor unsupervised pretraining. To verify the impact of introducing the adversarial augmentation, we focus on the improvements of GraphACL over DGI and GCN. For Metattack, we only evaluate the model's performance on the evasive task.

Datasets are summarized in the supplementary materials. The results in Table \ref{results} and \ref{meta-results} denote averaged classification accuracy and standard deviation over 5 random seeds. GACL is short for GraphACL.


We denote DGI and GraphACL as pre-trained methods, which are unsupervised pre-trained with only the graph and features and without any other information in downstream tasks. No previous study includes pre-trained methods in robustness experiments; however, we find it effective to defend unknown attacks with pretraining. Our main hypothesis is that the adversarial augmentations will help the model learn more robust representations, which is confirmed by comparing DGI and GraphACL. Thereby, we highlight the best results in DGI, GCL and GraphACL in bold, surpassing all other results except few special cases. 

\subsection{Robustness Evaluation under Netattack}

\begin{small}
\begin{table}
\centering

\caption{Classification Accuracy(\%) under Netattack over 5 random seeds}
\label{results}
\resizebox{\textwidth}{!}{
\begin{tabular}{ccccccc|cccc}
\hline
\multirow{2}{*}{}  & \multirow{2}{*}{\textbf{Model}} & \multirow{2}{*}{\textbf{Clean}} & \multicolumn{4}{c}{\textbf{Evasive}}         & \multicolumn{4}{c}{\textbf{Poisoning}}       \\ \cline{4-11}
                                   &                                 & \textbf{}          & \textbf{1} & \textbf{2} & \textbf{3} & \textbf{4} & \textbf{1} & \textbf{2} & \textbf{3} & \textbf{4} \\ \hline
\multirow{5}{*}{\rotatebox{90}{\textbf{Cora}}}     & \textbf{GCN}                    & 80.0\scriptsize{$\pm$7.87}          & 51.5\scriptsize{$\pm$4.87}  & 38.0\scriptsize{$\pm$6.22}  & 31.0\scriptsize{$\pm$2.24}  & 26.0\scriptsize{$\pm$3.79}  & 47.5\scriptsize{$\pm$7.07}  & 39.5\scriptsize{$\pm$2.74}  & 30.0\scriptsize{$\pm$5.00}  & 26.5\scriptsize{$\pm$3.79}  \\
                                   & \textbf{RGCN}                   & 80.0\scriptsize{$\pm$4.67}          & 49.5\scriptsize{$\pm$6.47}  & 36.0\scriptsize{$\pm$5.18}  & 30.5\scriptsize{$\pm$3.25}  & 25.5\scriptsize{$\pm$2.09}  & 46.5\scriptsize{$\pm$5.75}  & 35.5\scriptsize{$\pm$3.70}  & 29.0\scriptsize{$\pm$3.79}  & 25.5\scriptsize{$\pm$2.73}  \\
                                   & \textbf{GAT}                    & 77.8\scriptsize{$\pm$3.97}          & 48.0\scriptsize{$\pm$8.73}  & 39.5\scriptsize{$\pm$5.70}  & 36.5\scriptsize{$\pm$5.48}  & 32.5\scriptsize{$\pm$5.30}  & 50.5\scriptsize{$\pm$5.70}  & 38.0\scriptsize{$\pm$5.97}  & 33.5\scriptsize{$\pm$2.85}  & 26.0\scriptsize{$\pm$3.79}  \\
                                   & \textbf{DGI}               & 82.5\scriptsize{$\pm$4.33}          & 62.0\scriptsize{$\pm$4.81}  & 46.0\scriptsize{$\pm$3.79}  & 34.0\scriptsize{$\pm$5.18}  & {27.5}\scriptsize{$\pm$3.06}  & 62.5\scriptsize{$\pm$3.54}  & 43.5\scriptsize{$\pm$3.79}  & {31.5}\scriptsize{$\pm$6.75}  & 26.5\scriptsize{$\pm$4.18}  \\
                            & \textbf{GCL} & 64.4\scriptsize{$\pm$4.27} & 53.8\scriptsize{$\pm$5.20} & 38.8\scriptsize{$\pm$8.54} & 25.6\scriptsize{$\pm$3.15} & 18.8\scriptsize{$\pm$3.23} & 41.9\scriptsize{$\pm$6.25} & 33.1\scriptsize{$\pm$8.00} & 28.8\scriptsize{$\pm$7.77} & 23.1\scriptsize{$\pm$5.54} \\
                                   & \textbf{GACL}                & 82.0\scriptsize{$\pm$3.26}          & \textbf{67.5}\scriptsize{$\pm$5.00}  & \textbf{46.0}\scriptsize{$\pm$5.76}  & \textbf{35.5}\scriptsize{$\pm$4.81}  & \textbf{29.0}\scriptsize{$\pm$6.75}  & \textbf{64.5}\scriptsize{$\pm$5.70}  & \textbf{44.0}\scriptsize{$\pm$7.42}  & \textbf{33.5}\scriptsize{$\pm$6.02}  & \textbf{27.5}\scriptsize{$\pm$5.30}  \\ \hline
\multirow{5}{*}{\rotatebox{90}{\textbf{Citeseer}}} & \textbf{GCN}                    & 71.8\scriptsize{$\pm$6.94}          & 42.5\scriptsize{$\pm$7.07}  & 27.5\scriptsize{$\pm$6.37}  & 18.0\scriptsize{$\pm$3.26}  & 15.0\scriptsize{$\pm$2.50}  & 29.0\scriptsize{$\pm$7.20}  & 20.5\scriptsize{$\pm$1.12}  & 17.5\scriptsize{$\pm$1.77}  & 13.0\scriptsize{$\pm$2.09}  \\
                                   & \textbf{RGCN}                   & 73.5\scriptsize{$\pm$8.40}          & 41.5\scriptsize{$\pm$7.42}  & 24.5\scriptsize{$\pm$6.47}  & 18.5\scriptsize{$\pm$6.52}  & 13.0\scriptsize{$\pm$1.11}  & 31.0\scriptsize{$\pm$5.48}  & 19.5\scriptsize{$\pm$2.09}  & 13.5\scriptsize{$\pm$2.85}  & 5.00\scriptsize{$\pm$1.77}  \\
                                   & \textbf{GAT}                    & 72.3\scriptsize{$\pm$8.38}          & {49.0}\scriptsize{$\pm$9.12}  & 33.0\scriptsize{$\pm$5.97}  & 22.0\scriptsize{$\pm$4.81}  & 18.0\scriptsize{$\pm$3.26}  & 38.0\scriptsize{$\pm$5.12}  & 23.5\scriptsize{$\pm$4.87}  & 16.5\scriptsize{$\pm$4.54}  & 12.0\scriptsize{$\pm$2.09}  \\
                                   & \textbf{DGI}               & 78.5\scriptsize{$\pm$5.76}          & 64.0\scriptsize{$\pm$4.18}  & 49.5\scriptsize{$\pm$4.47}  & 36.5\scriptsize{$\pm$5.18}  & 30.5\scriptsize{$\pm$5.97}  & 57.5\scriptsize{$\pm$4.68}  & 40.0\scriptsize{$\pm$7.70}  & 31.0\scriptsize{$\pm$2.24}  & {25.5}\scriptsize{$\pm$5.70}  \\
                                    & \textbf{GCL} & 70.0\scriptsize{$\pm$7.36} & 59.4\scriptsize{$\pm$6.57} & 47.5\scriptsize{$\pm$5.40} & 36.3\scriptsize{$\pm$6.29} & 32.5\scriptsize{$\pm$4.08} & 50.6\scriptsize{$\pm$5.54} & 38.3\scriptsize{$\pm$6.77} & 38.1\scriptsize{$\pm$4.27} & 26.9\scriptsize{$\pm$8.00}\\
                                   & \textbf{GACL}               & 77.5\scriptsize{$\pm$3.06}          & \textbf{66.0}\scriptsize{$\pm$3.79}  & \textbf{53.0}\scriptsize{$\pm$5.70}  & \textbf{46.0}\scriptsize{$\pm$4.87}  & \textbf{37.0}\scriptsize{$\pm$1.12}  & \textbf{63.5}\scriptsize{$\pm$4.18}  & \textbf{41.0}\scriptsize{$\pm$3.79}  & \textbf{40.0}\scriptsize{$\pm$9.19}  & \textbf{30.5}\scriptsize{$\pm$5.97}  \\ \hline
\multirow{5}{*}{\rotatebox{90}{\textbf{Pubmed}}} & \textbf{GCN}                             & 82.6\scriptsize{$\pm$6.98}          & 39.5\scriptsize{$\pm$4.81}  & 32.0\scriptsize{$\pm$4.81}  & 31.0\scriptsize{$\pm$5.76}  & 31.0\scriptsize{$\pm$5.76}  & 36.0\scriptsize{$\pm$4.18}  & 32.5\scriptsize{$\pm$6.37}  & 31.0\scriptsize{$\pm$5.76}  & 28.5\scriptsize{$\pm$5.18}  \\
                                   & \textbf{RGCN}                   & 79.0\scriptsize{$\pm$5.18}          & 39.5\scriptsize{$\pm$5.70}  & 33.0\scriptsize{$\pm$4.80}  & 31.5\scriptsize{$\pm$4.18}  & 30.0\scriptsize{$\pm$5.00}  & 38.5\scriptsize{$\pm$4.18}  & 31.5\scriptsize{$\pm$2.85}  & 29.5\scriptsize{$\pm$3.70}  & 27.0\scriptsize{$\pm$3.70}  \\
                                   & \textbf{GAT}                    & 78.6\scriptsize{$\pm$6.70}          & 41.0\scriptsize{$\pm$8.40}  & 33.5\scriptsize{$\pm$4.18}  & 30.5\scriptsize{$\pm$4.47}  & 31.0\scriptsize{$\pm$4.18}  & 39.5\scriptsize{$\pm$3.26}  & 31.0\scriptsize{$\pm$4.18}  & 30.0\scriptsize{$\pm$3.06}  & 35.5\scriptsize{$\pm$5.97}  \\
                                   & \textbf{DGI}               & 79.0\scriptsize{$\pm$7.20}          & 40.5\scriptsize{$\pm$5.86}  & 31.0\scriptsize{$\pm$4.54}  & 29.5\scriptsize{$\pm$3.71}  & 28.0\scriptsize{$\pm$2.74}  & 40.0\scriptsize{$\pm$4.81}  & 31.0\scriptsize{$\pm$3.79}  & 28.5\scriptsize{$\pm$4.18}  & 28.0\scriptsize{$\pm$4.68}  \\
                                    & \textbf{GCL} & 67.5\scriptsize{$\pm$7.07} & \textbf{45.3}\scriptsize{$\pm$1.77} & \textbf{35.8}\scriptsize{$\pm$6.29} & 28.3\scriptsize{$\pm$5.10} & 28.1\scriptsize{$\pm$4.26} & 40.25\scriptsize{$\pm$1.77} & 33.25\scriptsize{$\pm$6.61} & \textbf{30.3}\scriptsize{$\pm$1.44} & 19.4\scriptsize{$\pm$3.75}\\
                                   & \textbf{GACL}               & 83.0\scriptsize{$\pm$5.42}          & 43.0\scriptsize{$\pm$5.42}  & 34.0\scriptsize{$\pm$5.18}  & \textbf{30.0}\scriptsize{$\pm$3.06}  & \textbf{28.5}\scriptsize{$\pm$4.18}  & \textbf{41.0}\scriptsize{$\pm$3.79}  & \textbf{34.0}\scriptsize{$\pm$2.85}  & 29.5\scriptsize{$\pm$3.71}  & \textbf{28.5}\scriptsize{$\pm$4.18}  \\ \hline
\end{tabular}
}
\end{table}
\end{small}

\textbf{Netattack.}
For experiments on Cora, GraphACL with pretraining and adversarial augmentations, outperforms all previous methods like GCN remarkably. In evasive experiments, GraphACL surpasses GCN by 16.0\%, RGCN by 18.0\%, GAT by 19.5\% and DGI by 5.5\% on average on the task with one perturbation. Also, GraphACL achieves 17.0\% and 2.0\% improvements over GCN and DGI respectively, when being poisoned on one perturbation case. When the number of perturbations gets larger, the averaged results of GraphACL are still a bit higher.

GraphACL also achieves significant improvements on both evasive and poisoning experiments on Citeseer. Note that many nodes in Citeseer have few degrees, thus making the attack much harder to defend. When the number of perturbations is 1, GraphACL surpasses DGI and GCN by 2.0\% and 23.5\% on evasive tasks, 3.5\% and 34.5\% on poisoning tasks. Especially, when the number of perturbations is 3, GraphACL surpasses DGI by 9.5\% on evasive tasks and 9.0\% on poisoning tasks. All results of GraphACL surpass DGI and GCN a lot on both evasive and poisoning tasks. We attribute the success to the added views of adversarial augmentations, which makes the model more defensive to the unseen attacks on graphs like Citeseer. 

Additionally, GraphACL achieves the best on the clean graph and obtains similar results to GraphCL on Pubmed both in evasive setting and poisoning setting. When the number of perturbations is 1, GraphACL improves the averaged accuracy on GCN, RGCN, GAT, and DGI by 3.5\%, 3.5\%, 2.0\%, and 2.5\%, respectively on evasive tasks.


\subsection{Robustness Evaluation under Metattack}
\textbf{Metattack.}
We also evaluate the performances of DGI and GraphACL on Cora and Citeseer under Metattack in Table~\ref{meta-results}. We use our proposed unsupervised attack method in Figure~\ref{fig:attack_unsu} to generate graph adversarial samples for GraphACL. On the second row, clean denotes evaluation on the clean graph after Metattack, while 0.05 and 0.2 denotes the perturbed rate on the graph. The first column denotes the the perturbed rate in GraphACL training, which is related to adversarial generation. When this rate is 0.000, the method is indeed DGI. While the rate increases, the perturbation gets stronger. As information bottleneck demonstrated, there would be a desired representation containing sufficient information with less nuisance to get more robust performance. The results in Table \ref{meta-results} prove the same idea. The performance gets higher first, achieves a peak, and then goes down when the perturbed rate increases. Surprisingly, we find that the optimal results are all related to GraphACL with 0.030 rate of perturbation. If the perturbation rate is much larger, the graph is corrupted too much to maintain sufficient information, thus resulting in poor performance. The best performances of GraphACL all outperform DGI. When the evalutaion graph is more perturbed, which means the robust representation is much more needed, the improvements get much higher. In Cora, the best GraphACL achieves 1.4\%, 2.2\% and 3.5\% higher in performance than DGI on clean, 0.05-perturbed, 0.2-perturbed graph respectively. In Citeseer, the best GraphACL achieves 2.0\%, 5.4\% and 3.2\% higher in performance than DGI on clean, 0.05-perturbed, 0.2-perturbed graph respectively.

\begin{table*}[hbtp]
\centering
\caption{Classification Accuracy(\%) under Metattack over 5 random seeds}
\begin{tabular}{ccccccc}
\hline
\textbf{Dataset}                        & \multicolumn{3}{c}{\textbf{Cora}}                          & \multicolumn{3}{c}{\textbf{Citeseer}}                      \\ \hline
\textbf{Rate \textbackslash{}Task} & \textbf{Clean}       & \textbf{0.05}        & \textbf{0.2} & \textbf{Clean}       & \textbf{0.05}        & \textbf{0.2} \\ \hline
\textbf{DGI}                            & 75.2\scriptsize{$\pm$2.71}          & 73.8\scriptsize{$\pm$2.48}          &        71.5\scriptsize{$\pm$2.77}      & 67.5\scriptsize{$\pm$3.44}          &  64.9\scriptsize{$\pm$4.18}          & 65.9\scriptsize{$\pm$4.10} \\
\textbf{GraphACL-0.001}                            & 75.4\scriptsize{$\pm$2.71}          & 74.7\scriptsize{$\pm$3.23}          &        73.1\scriptsize{$\pm$3.56}      & 69.1\scriptsize{$\pm$1.47}          &  69.0\scriptsize{$\pm$1.56}          & 68.9\scriptsize{$\pm$1.63} \\
\textbf{GraphACL-0.010}                            & 76.6\scriptsize{$\pm$2.21}          & 75.7\scriptsize{$\pm$2.47}          &        74.1\scriptsize{$\pm$2.55}      & 70.2\scriptsize{$\pm$1.90}          &  69.9\scriptsize{$\pm$2.01}          & 68.8\scriptsize{$\pm$2.04} \\
\textbf{GraphACL-0.020}                            & 76.2\scriptsize{$\pm$2.22}          & 74.8\scriptsize{$\pm$2.93}          &        74.9\scriptsize{$\pm$3.84}      & 68.6\scriptsize{$\pm$3.59}          &  68.0\scriptsize{$\pm$3.83}          & 66.4\scriptsize{$\pm$3.28} \\
\textbf{GraphACL-0.030}                            & \textbf{76.8}\scriptsize{$\pm$0.82}          & \textbf{76.0}\scriptsize{$\pm$1.40}          &        \textbf{75.0}\scriptsize{$\pm$1.70}      & \textbf{70.5}\scriptsize{$\pm$1.88}          &  \textbf{70.3}\scriptsize{$\pm$1.87}          & \textbf{69.1}\scriptsize{$\pm$2.40} \\
\textbf{GraphACL-0.040}                            & 73.7\scriptsize{$\pm$3.95}          & 73.3\scriptsize{$\pm$3.90}          &        72.4\scriptsize{$\pm$3.96}      & 69.2\scriptsize{$\pm$2.50}          &  68.8\scriptsize{$\pm$2.54}          & 67.3\scriptsize{$\pm$3.21} \\
\textbf{GraphACL-0.050}                            & 74.4\scriptsize{$\pm$4.11}          & 73.8\scriptsize{$\pm$4.67}          &        72.8\scriptsize{$\pm$5.55}      & 69.8\scriptsize{$\pm$1.70}          &  69.5\scriptsize{$\pm$1.88}          & 68.2\scriptsize{$\pm$2.35} \\ \hline
\end{tabular}
\label{meta-results}
\end{table*}

\subsection{Perturbation Rate Sensitivity for Adversarial Samples}

\begin{figure*}[htbp]
  \centering
  \subfigure{\includegraphics[width=0.333\textwidth]{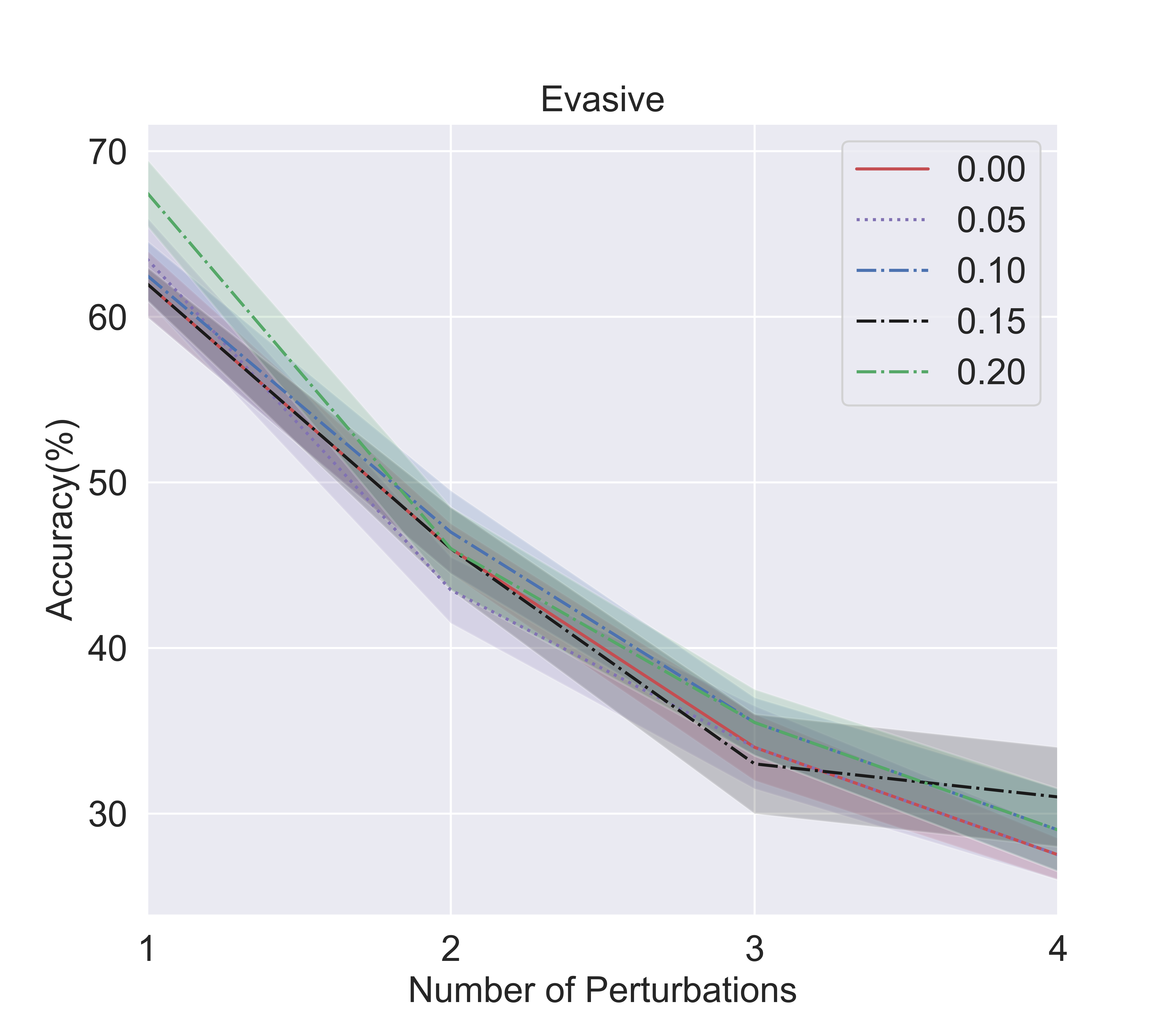}}
  \subfigure{\includegraphics[width=0.333\textwidth]{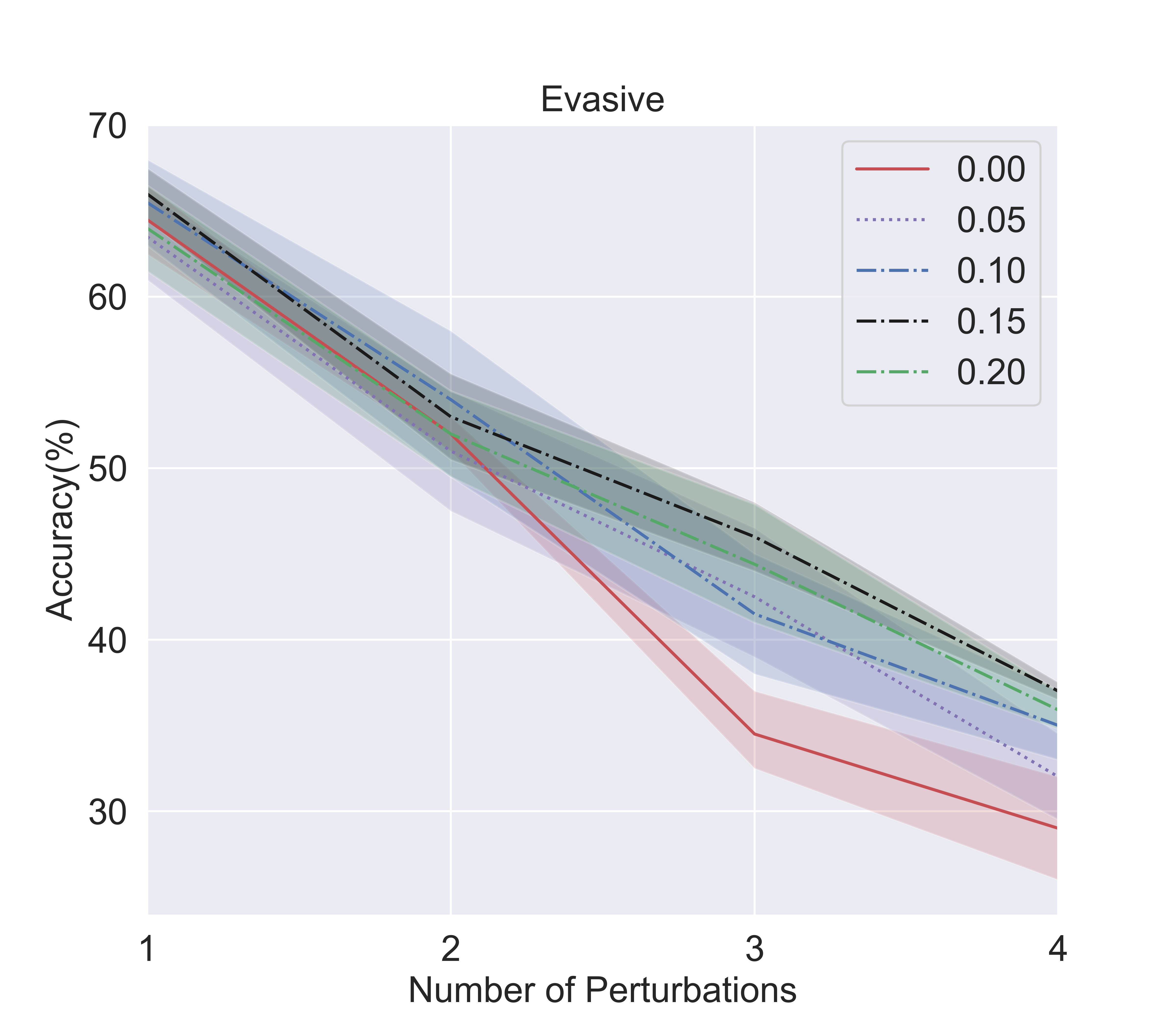}}
    \subfigure{\includegraphics[width=0.309\textwidth]{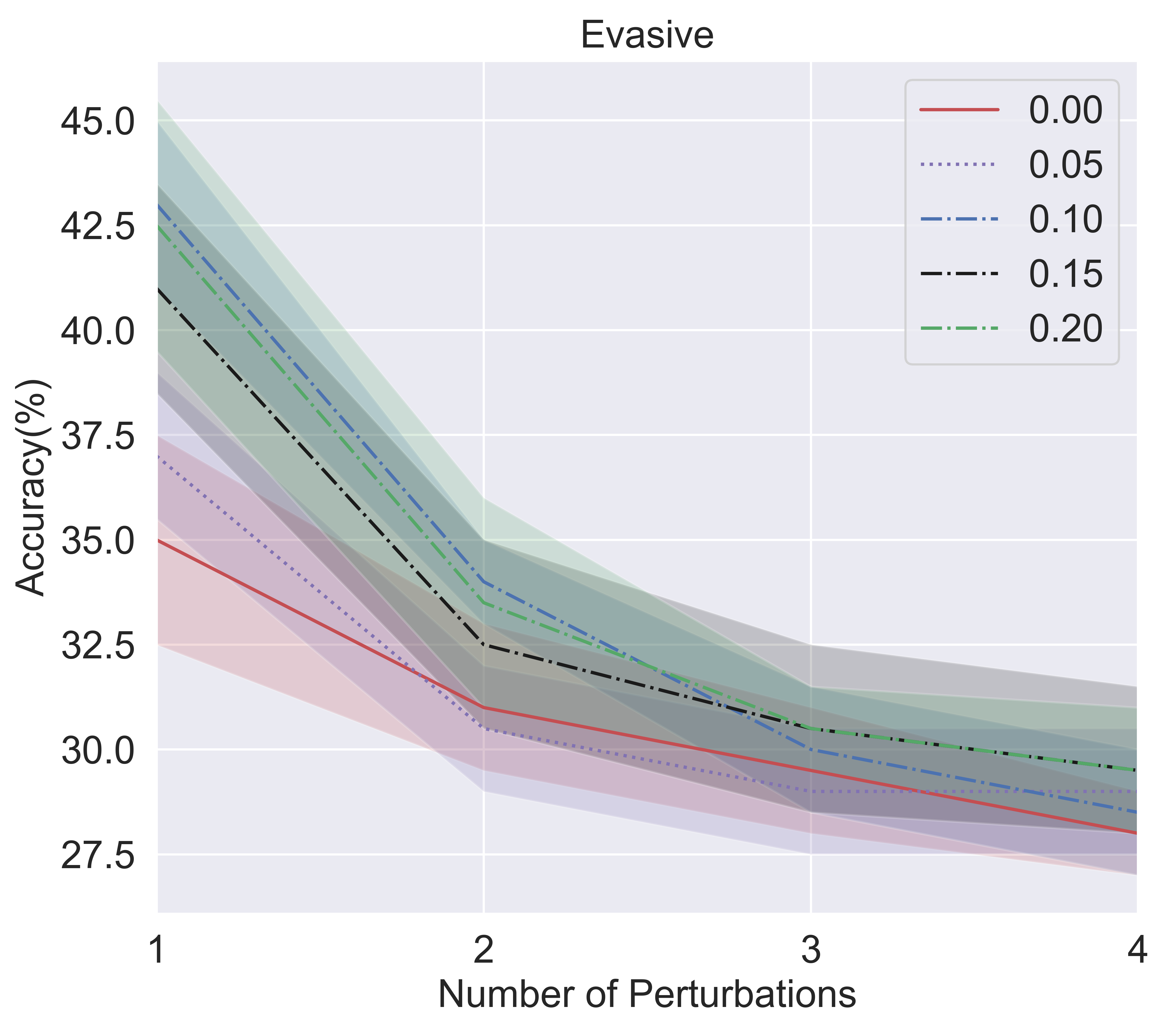}}
  \subfigure{\includegraphics[width=0.333\textwidth]{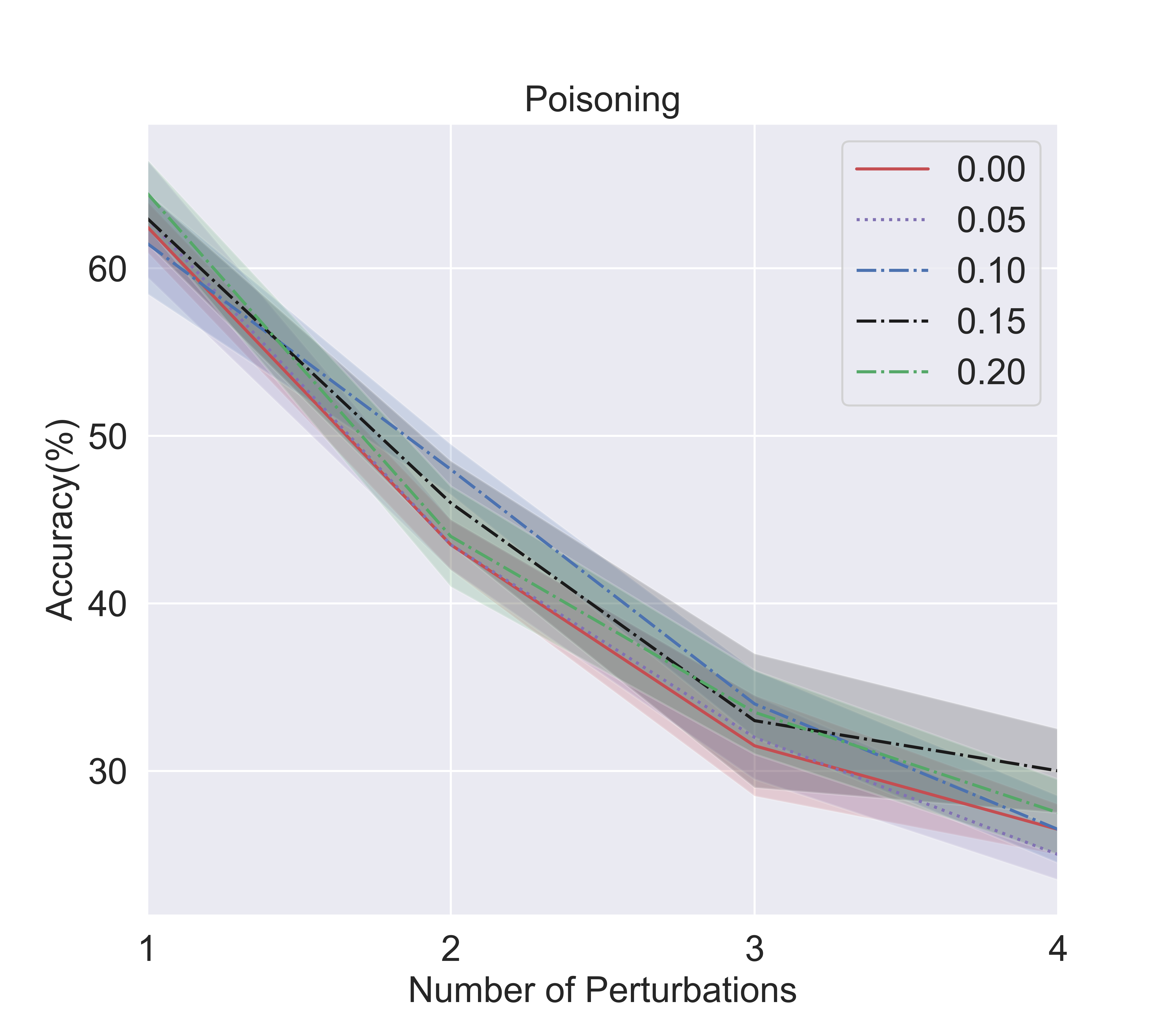}}
  \subfigure{\includegraphics[width=0.333\textwidth]{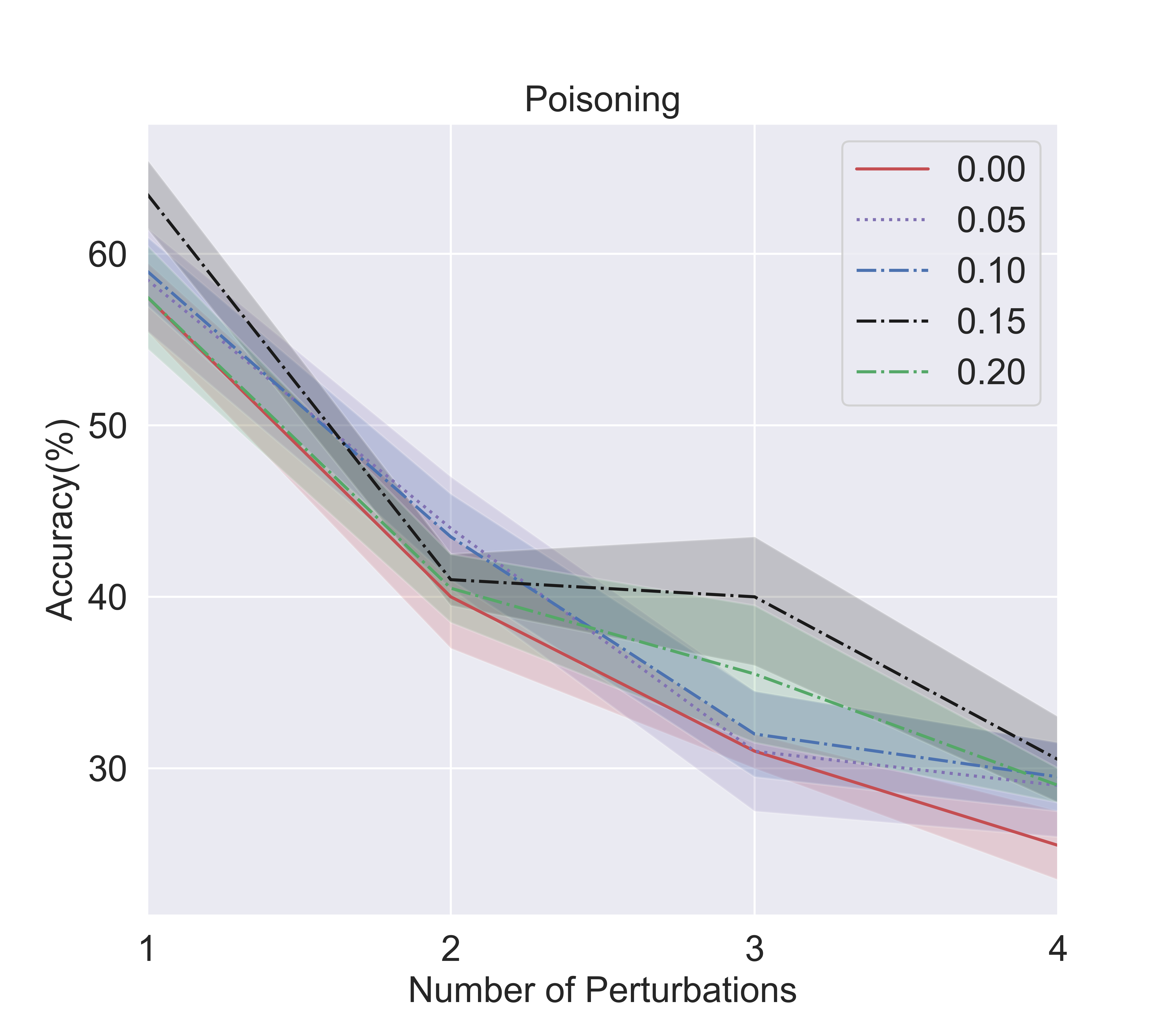}}
  \subfigure{\includegraphics[width=0.309\textwidth]{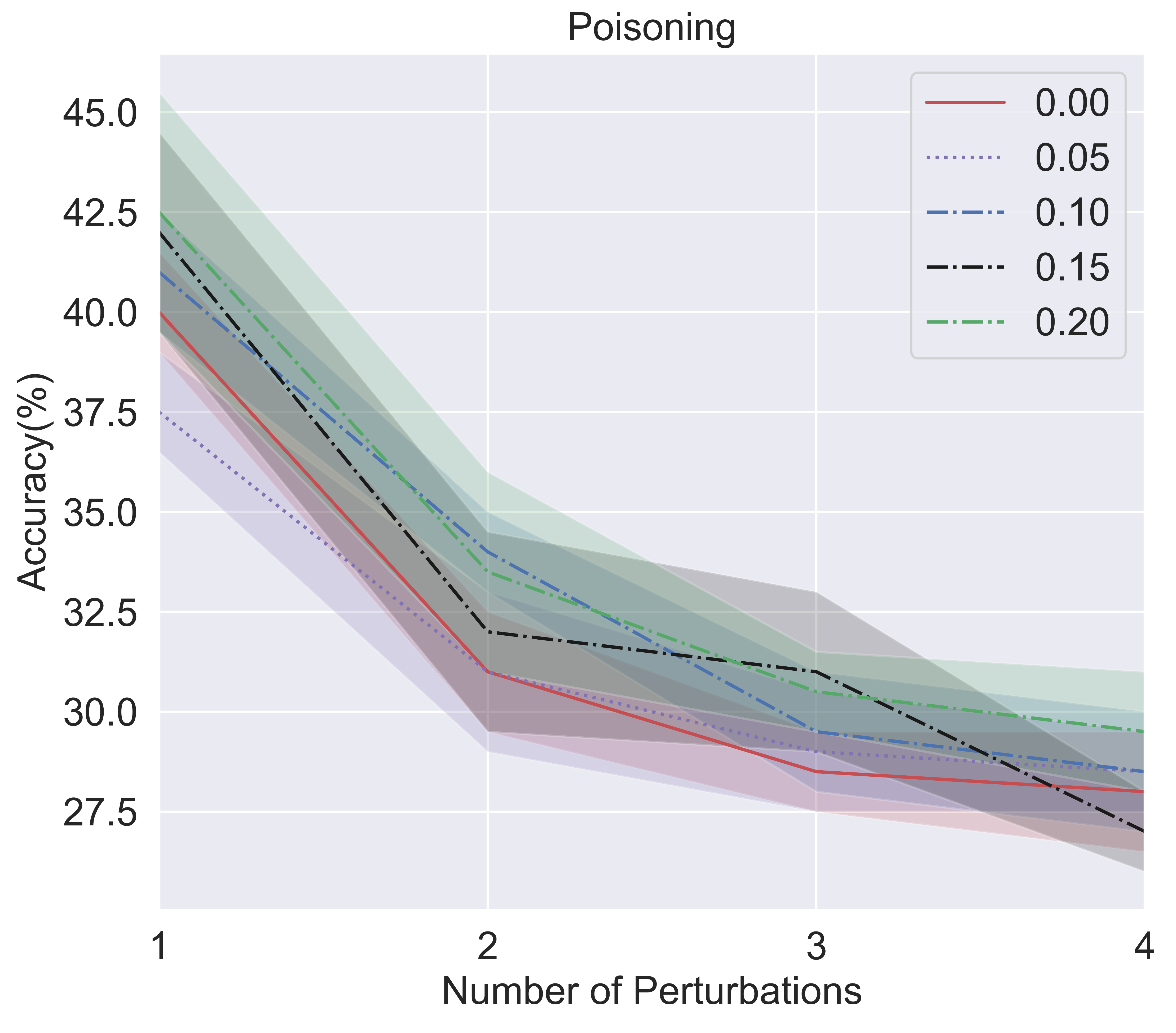}}
  \caption{Results on different perturbation rates for adversarial samples}
  \label{pert_rate}
\end{figure*}

In Figure \ref{pert_rate}, we evaluate GraphACL with different perturbation rates for adversarial samples on Cora, Citeseer and Pubmed. Experiments are conducted over 5 random seeds. The solid line denotes GraphACL with zero perturbation rate served as a baseline, which means no augmentation is included, i.e., DGI. The dotted line denotes GraphACL with the least positive perturbation rate, which usually performs the worst within GraphACL. The other three lines in dash-dot style related to GraphACL with different suitable perturbation rates. Their performances are similar, which means our method is not sensitive to the perturbation rate in a reasonable range. The difference is mainly based on the dataset. Within the range, there is a best perturbation rate for GraphACL to improve the performance over baselines up to 8\%.

\section{Related work}

\subsection{Adversarial Attack and Defense on Graph Data}
The first graph adversarial attack is proposed by Zugner et al. to generate adversarial graph data using an efficient greedy search method ~\cite{10.1145/3219819.3220078}. The generated graph can be used to fool GNN or other traditional graph learning methods. Then some methods are proposed to attack the topological structure by adding or removing edges according to the gradient of a surrogate model. Xu et al. proposed an optimization-based attack method based on the gradient of the surrogate model~\cite{ijcai2019-550}. Zugner et al. presented to use meta-gradient to guide the perturbation of graph adjacency matrix ~\cite{zugner_adversarial_2019}. Wu et al. argued that integrated gradients can better reflect the effect of perturbing certain features or edges ~\cite{ijcai2019-669}. Also, Bojchevski et al. took DeepWalk~\cite{10.1145/2623330.2623732} as base method using eigen-decomposition and genetic algorithm based strategy to attack the network embedding ~\cite{pmlr-v97-bojchevski19a}.

To ensure the robustness under adversarial attack, many methods have been proposed to defense GNN models~\cite{wang2019adversarial,DBLP:journals/corr/abs-1903-05994,8924766,DBLP:journals/corr/abs-1911-04429,zhu2019robust,wu2020graph}. Wang et al. thought the vulnerabilities of graph neural networks are related to the aggregation layer and the perceptron layer ~\cite{wang2019adversarial}. To address these two disadvantages, they propose an adversarial training framework with a modified GNN model to improve the robustness of GNNs.  Chen et al. proposed different defense strategies based on adversarial training for target and global adversarial attack with smoothing distillation and smoothing cross-entropy loss function \cite{DBLP:journals/corr/abs-1903-05994}. Feng et al. proposed a method of adversarial training for the attack on node features with a graph adversarial regularizer which encourages the model to
generate similar predictions on the perturbed target node and its connected nodes ~\cite{8924766}. Wang et al. pointed out that the values of perturbation could be continuous or even negative ~\cite{DBLP:journals/corr/abs-1911-04429}. Zhu et al. proposed to use Gaussian distribution to increase the robustness of Graph Convolutional Network ~\cite{zhu2019robust}. Wu et al. applied the information bottleneck principle on semi-supervised learning settings~\cite{wu2020graph} to defense the targeted node attack~\cite{10.1145/3219819.3220078}.
\subsection{Self-supervised Graph Representation Learning}

Self-supervised contrastive learning~\cite{pmlr-v119-chen20j,He_2020_CVPR} showed significant performance on graph-structured data. The contrastive learning approach usually needs to generate augmented graph sample pairs of the original graph. Then the similarity between the representation of augmented graph pairs is minimized to learn graph representations. Veli{\v{c}}kovi{\'c} et al. proposed to maximize the information between local and global graph representations to learn node representations~\cite{velivckovic2018deep}. Zhu et al. proposed various augmentation strategies to generate augmented graph samples ~\cite{zhu2020deep}. Hassani et al. introduced multi-view contrastive learning ~\cite{pmlr-v119-hassani20a} that maximizes the information between graph and its diffusion versions~\cite{NEURIPS2019_23c89427}. Qiu et al. used an anonymous random walk to generate augmented subgraphs from a large graph and minimize the similarity between the paired subgraphs and maximize the similarity between subgraphs and negative samples ~\cite{10.1145/3394486.3403168}. Although various methods have been proposed to use self-supervised contrastive learning to learn graph representations, few works
considered the quality of augmentation samples.

\section{Conclusion and Discussion}

To summarize, we introduce adversarial augmentations into graph self-supervised representation learning and propose a novel Graph Adversarial Contrastive Learning (GraphACL) framework. Theoretically, we obtain an upper bound of the Information Bottleneck loss function for graph contrastive learning. With adversarial augmentations, our method could result in a much tighter bound and more robust representations. Based on the theoretical analysis, we formulate the GraphACL framework and present relative objective functions in both supervised and unsupervised settings. To verify the empirical performance, we conduct experiments on classic benchmarks attacked by Netattack or Metattack. GraphACL outperforms DGI and other baselines on both evasive and poisoning tasks, thus proving itself a more robust way of graph representation learning. The analysis of different perturbation rates also indicates that our method is not sensitive to the rate. Albeit our model is built on top of Deep graph infomax (DGI), our theory can be easily extended to other models by combining adversarial learning and graph self-supervised learning together.
%
%
\bibliographystyle{splncs04}
\bibliography{main}
%

%
%
%
%
\end{document}